\documentclass[letterpaper]{article} 
\usepackage{aaai25}  
\usepackage{times}  
\usepackage{helvet}  
\usepackage{courier}  
\usepackage[hyphens]{url}  
\usepackage{graphicx} 
\usepackage{natbib}  
\usepackage{caption} 
\usepackage{algorithm}
\usepackage{algorithmic}
\usepackage{amsmath} 
\usepackage{amssymb}
\usepackage{enumitem}
\usepackage{pifont}

\usepackage{newfloat}
\usepackage{listings}
\DeclareCaptionStyle{ruled}{labelfont=normalfont,labelsep=colon,strut=off} 
\floatstyle{ruled}
\newfloat{listing}{tb}{lst}{}
\floatname{listing}{Listing}
\title{MTGA: Multi-View Temporal Granularity Aligned Aggregation for Event-Based Lip-Reading}

\author{
Wenhao Zhang\textsuperscript{\rm 1}\equalcontrib,
Jun Wang\textsuperscript{\rm 1}\equalcontrib,
Yong Luo\textsuperscript{\rm 1},
Lei Yu\textsuperscript{\rm 2},
Wei Yu\textsuperscript{\rm 1}, 
Zheng He\textsuperscript{\rm 1}\thanks{Corresponding author.}, 
Jialie Shen\textsuperscript{\rm 3}
}

\affiliations{
    \textsuperscript{\rm 1}School of Computer Science, National Engineering Research Center for Multimedia Software and Hubei Key Laboratory of Multimedia and Network Communication Engineering, Wuhan University, China\\

    \textsuperscript{\rm 2}School of Electronics and Information, Wuhan University, China\\
    \textsuperscript{\rm 3}Department of Computer Science, City, University of London, The United Kingdom\\
    \{wenhaozhang, junwang\_ai, luoyong, ly.wd, yuwei, hezheng\}@whu.edu.cn, jialie@gmail.com
}

\usepackage{bibentry}

\begin{document}

\maketitle

\begin{abstract}
Lip-reading is to utilize the visual information of the speaker’s lip movements to recognize words and sentences. Existing event-based lip-reading solutions integrate different frame rate branches to learn spatio-temporal features of varying granularities. However, aggregating events into event frames inevitably leads to the loss of fine-grained temporal information within frames. To remedy this drawback, we propose a novel framework termed Multi-view Temporal Granularity aligned Aggregation (MTGA). Specifically, we first present a novel event representation method, namely time-segmented voxel graph list, where the most significant local voxels are temporally connected into a graph list. Then we design a spatio-temporal fusion module based on temporal granularity alignment, where the global spatial features extracted from event frames, together with the local relative spatial and temporal features contained in voxel graph list are effectively aligned and integrated. Finally, we design a temporal aggregation module that incorporates positional encoding, which enables the capture of local absolute spatial and global temporal information. Experiments demonstrate that our method outperforms both the event-based and video-based lip-reading counterparts. 
\end{abstract}


%
\begin{links}
   \link{Code}{https://github.com/whu125/MTGA}
\end{links}

\section{Introduction}
With the continuous advancement of human-computer interaction technology, lip reading as a silent form of communication has attracted widespread attention. Existing lip-reading approaches are typically based on video inputs \cite{sheng2024deep}, which are required to be high resolution. Poor performance may be obtained when the video is blurry or involves rapid movement.

Event cameras \cite{gallego2020event}, as innovative visual sensors, can capture changes in pixel brightness at the microsecond level, producing an event stream output. Their sensitivity to motion and ability to reduce redundancy make them well-suited for lip-reading tasks that require capturing fine-grained features and lip movements.

There exist many event-based recognition approaches. For example, the event streams are represented as point clouds in \cite{wang2019space} to preserve maximal information, but the computational cost may be very high for a large number of events.
Li \textit{et al.} ~\cite{li2021graph} employ neighboring graph structures to capture of local correlations, yet lip-reading task may introduce an excessive number of nodes and edges, and thus make the feature extraction difficult. Some other approaches assemble asynchronous events into group tokens \cite{peng2023get}, based on timestamps and polarity. This can efficiently reduce feature dimension but do not adequately exploit the temporal information, rendering them less effective for lip-reading task that demands a focus on subtle temporal variations.

A recent lip-reading approach based on event camera is presented in \cite{tan2022multi}.
In this approach, the event points within a set of time are aggregated into a single frame, so that the video processing strategies can be employed. 
However, during the talking, lips often change subtly and rapidly, and using image frame for representation may obscure these detailed variations. Tan \textit{et al.} ~\cite{tan2022multi} alleviate this issue by developing a Multi-grained Spatio-Temporal Features Perceived (MSTP) Network that merges features learned from two branches with different frame rates through a Message Flow Module (MFM). However, MSTP only aggregate the events into frames, and hence inevitably leads to the loss of intra-frame local information.

\begin{figure}
  \includegraphics[width=\linewidth]{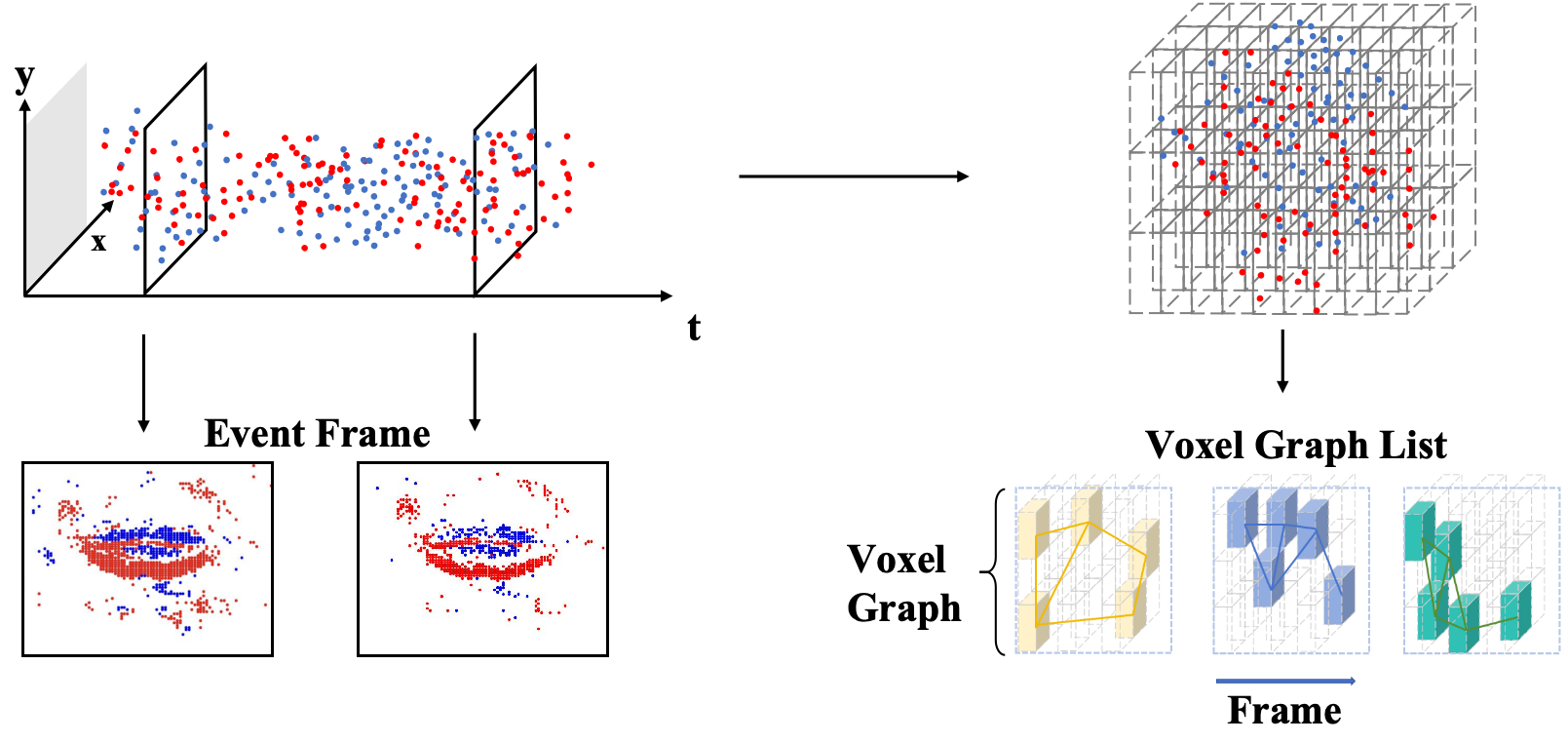}
  \caption{The two event representations we adopt. (Left) An event frame is obtained by integrating event points within a certain temporal range in the event stream. (Right) After dividing the event stream into a voxel grid in three-dimensional space, a three-dimensional geomeric neighboring graph is constructed at regular time intervals, resulting in a voxel graph list used for event representation.} \label{fig1}
\end{figure}

To remedy this drawback, we propose a novel multi-view learning method termed multi-view temporal granularity aligned aggregation (MTGA) for event-based lip-reading. In particular, we first represent the events from two different viewpoints, as shown in Figure~\ref{fig1}. One view is to aggregate the events into event frames at an appropriate temporal segment following~\cite{tan2022multi}, and convolution is applied to extract global spatial features. The other view is to partition the event flow (at the same temporal segment as the first view) into a voxel grid within a three-dimensional space. Within each time segment, we construct a three-dimensional geometric neighboring graph using the most informative voxels. The resulting local graphs are temporally connected to form a graph list, which are utilized to extract critical local details and contextual correlations using Gaussian Mixture Model convolutions~\cite{li2021graph}.

Our designed voxel graph list are temporally aligned with the event frames, and then we devise a fusion module that combines features extracted from both branches through convolution within each temporal segment. Due to the stronger spatio-temporal connectivity, the voxel graphs can serve as supplement to the global spatial information extracted from the event frame, and thereby result in more discriminative spatio-temporal features.

Finally, we propose a temporal aggregation module to combine the obtained features of different temporal segments. This is achieved by introducing the idea of positional encoding adopted in Transformer \cite{vaswani2017attention}.
When constructing the voxel graph, the absolute position information of the voxel in the events is discarded.
Therefore, to capture the local absolute spatial information, the voxel coordinates are utilized to obtain absolute positional features through positional encoding. The result embeddings are appended to the temporal segment features and fed into the Bi-GRU and Self-Attention layers to exploit the global temporal information.
We verify the effectiveness of our proposed method on the DVS-Lip \cite{tan2022multi} dataset, and the experiments demonstrate that our model can significantly outperform the state-of-the-arts in the field of event-based lip-reading recognition. For example, we obtain a $4.1\%$ relative improvement in terms of overall accuracy compared the most competitive counterpart.
In addition, we conducted experiments on the DVS128-Gait-Day dataset, and the experimental results proved that our model has good generalization performance.

In summary, the main contributions of our work are as follows:
\begin{itemize}
\item[$\bullet$] We propose a novel representation method for event streams termed temporal segmented voxel graph list, which can effectively exploit the local spatio-temporal correlations.
\item[$\bullet$] We design a temporal granularity aligned fusion module, which combines the voxel graph list features and frame features within each temporal segment, thereby obtaining more discriminative spatio-temporal features. 
\item[$\bullet$] We design a temporal aggregation module, which can capture the local position and global temporal information.
\item[$\bullet$] To the best of our knowledge, MTGA is the first work in the field of event-based lip reading that combines features from multiple views.
\end{itemize}
\section{Related Work}

\subsection{Lip-reading}
Lip-reading aims to understand what the people say according to the lip and facial movements.
Traditional lip-reading approaches extract handcrafted features and adopt classic classifiers for words recognition~\cite{anderson2013expressive,kim2015decision}.
Recently, some deep learning-based recognition networks are proposed for lip-reading.
Front-end of the network is used for feature extraction and some commonly utilized backbones are
CNNs \cite{li2021survey,sheng2021cross,feng2020learn}, GCNs \cite{sheng2021adaptive,liu2020lip,zhang2021lip}, and Visual Transformers \cite{prajwal2022sub,wang2019multi,han2022survey}. In the back-end, 
RNNs \cite{luo2020pseudo,zhao2020mutual,xiao2020deformation}, Transformer \cite{afouras2018deep,zhang2019spatio,vaswani2017attention}, and TCN \cite{martinez2020lipreading,afouras2018deep} are usually adopted to aggregate temporal information. For instance, Feng \cite{feng2020learn} \textit{et al.} chose the typical RNN-based Bi-GRU as the back-end network for lip-reading, where Bi-GRU's strong context learning and sequence modeling capabilities are effectively leveraged for word recognition. Martinez \cite{martinez2020lipreading} \textit{et al.} proposed a Multi-Scale Temporal Convolutional Network (MS-TCN) structure, which is able to capture long-term dependency.
Most of the existing lip recognition approaches are video-based. These approaches are limited in that the video resolution may be not enough to capture some subtle movements and the videos may contain motion blur. Inspired by~\cite{tan2022multi}, we proposed a method based on event streams, which often contains more rich information than videos.

\subsection{Event-based Recognition}
Event camera is a groundbreaking visual sensor,
where each pixel independently detects changes in luminance instead of capturing fixed-interval full-image frames \cite{gallego2020event}. When the brightness change of a pixel exceeds a preset threshold, the event camera generates an event point that includes the pixel's location, timestamp, and the polarity indicating the change in brightness. This allows for the capture of visual information at a high temporal resolution and provides a broad dynamic range. In the lip-reading application, high temporal resolution of event cameras is crucial for detecting subtle motion, and the sensitivity to motion makes them to particularly suitable to capture lip movements.

In event-based recognition, a representative work is presented in \cite{zhu2019unsupervised}, where the event stream is segmented by timestamps, and the event points within a certain timestamp range are unified to induce a single time point, resulting in 'event frames' similar to image frames.
The event streams can also be regarded as point clouds, and a representative work is conducted by Wang \cite{wang2019space} \textit{et al.}, which utilized PointNet for gesture recognition.
Another solution for event-based recognition is to integrate time and polarity information with tokens to represent events~\cite{peng2023get}, and this facilitates effective feature communication and integration in the spatial and temporal-polarity domains.
In~\cite{jiang2023point}, the events are represented as both point clouds and voxels, which are then constructed as graphs for feature extraction and combination using a graph network.
These event-based recognition approaches are used for the recognition of static objects or actions of short duration, and the temporal information are not exploited and thus cannot achieve satisfactory performance in the lip-reading task during a long time range.
As far as we know, the first lip-reading work based on event data is conducted by Tan \textit{et al.}~\cite{tan2022multi}, where a multi-granularity spatio-temporal feature perception network is designed to exploit the temporal information to some extent.
However, the events are only represented as event frames in~\cite{tan2022multi}, and thus the fine-grained information within frames cannot be exploited. This issue is addressed in this paper by integrating information from mulitple views.

\begin{figure*}[t!]
\includegraphics[width=\textwidth,trim={0 0 0 0},clip]{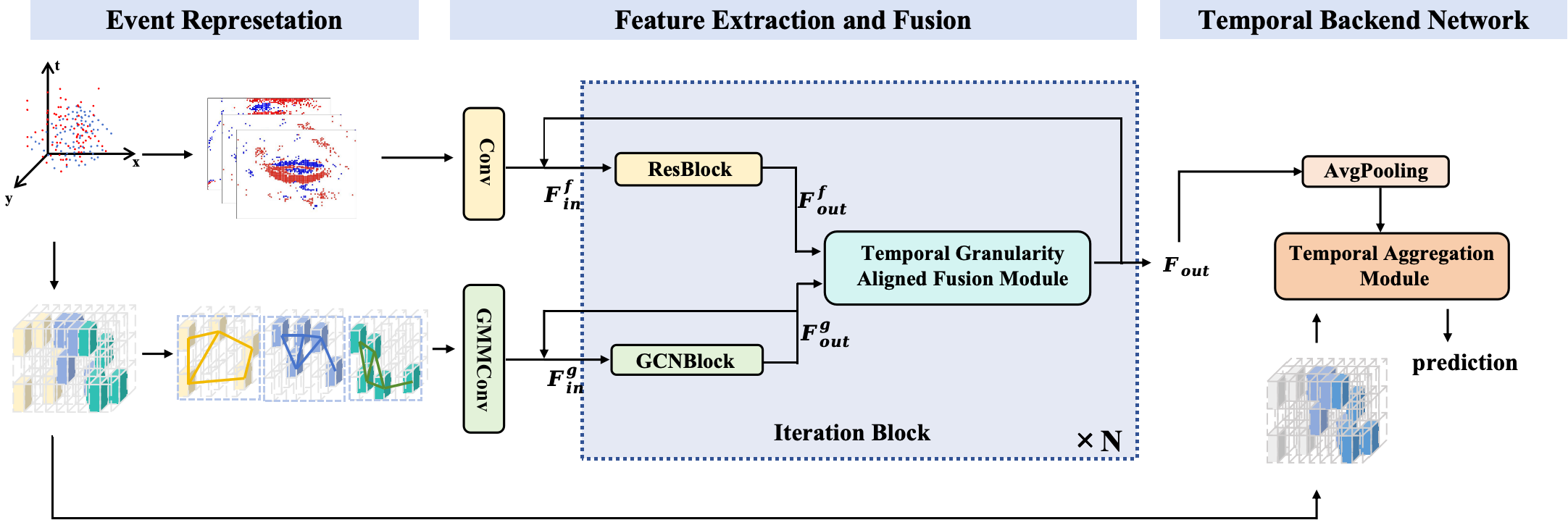}
\caption{The architecture of our proposed network. The model is divided into three components: (1) Event representation, which describes the events from two different viewpoints, i.e., event frames and voxel graph list; (2) Feature extraction and fusion, which extracts features for the two views separately, and combines them through a temporal granularity aligned fusion module; (3) Temporal backend network, which aggregates the global temporal information.
} \label{fig2}
\end{figure*}

\section{Methodology}
\subsection{Framework Overview}
In this paper, we present a multi-view learning method for event-based lip-reading, and an overview of the proposed framework is shown in Figure~\ref{fig2}. 
In the following, we will depict each module. 

\subsection{Multi-view Event Representation}
\subsubsection{Event Frame.} Converting the event stream into frames requires an intermediary, and a suitable choice is to encode the event stream into a spatio-temporal voxel grid \cite{zhu2019unsupervised}. For the event camera output, the event stream $\varepsilon = \{{(x_n, y_n, t_n, p_n)}\}^N_{n=1}$ includes $x$, $y$ direction coordinate, time and polarity. we discrete the timestamps of the event stream into $T$ time intervals through normalization. Each event point weights its polarity according to temporal proximity and is contributed to the two closest time intervals. The event frames $V(t,x,y)$ are generated using the method described in \cite{tan2022multi}.In our work, we divide the event stream into 60 event frames to represent the event stream.

\subsubsection{Voxel Graph List.} Another representation method initially maps the event data onto discretized voxels in a three-dimensional space. This approach enhances the capture of the events' spatial positioning and distribution information. To aggregate the event points while preserving polarity features, we convert the voxels into graph nodes, with the event points' polarities serving as the feature matrix of the nodes.

To align with event frames in the temporal dimension and supplement the temporal information within each frame, we first normalize time over $n*T$ (n=3 in our work), dividing the three-dimensional space sized $(n*T, H, W)$ into voxels with each voxel sized $(1, h, w)$, where $H$ and $W$ denote the spatial capture range of the event camera. . The three-dimensional space is respectively partitioned into $(n*T, \dfrac{H}{h}$, $\dfrac{W}{w})$ segments. After ensuring temporality aligned, each event frame corresponds to $n*\dfrac{H}{h}$*$\dfrac{W}{w}$ voxels. 


To reduce computational load, we select the k voxels with the highest number of event points per frame as graph nodes, denoting $\text{O}_i \in (\text{o}_{i1}...\text{o}_{ik})$ as the chosen voxels. We select $K$ event points within each voxel to obtain their polarity matrix as node features $\textbf{a} \in \text{R}^K$, thus each graph node is described by $\text{o}_{ij} \in (\text{t}_{ij},\text{x}_{ij},\text{j}_{ij},\textbf{a}_{ij})$. An edge exists between nodes when the Euclidean distance between their three-dimensional coordinates $[t_i,x_i,y_i]$ is less than threshold R, using this Euclidean distance as the edge feature, and the edge set is defined as $\text{E}_i$. The edge feature set $\text{W}_i \in \text{R}^{num\, edges}$ is obtained by taking the Euclidean distance as the feature of the edges. Represent the $i-$th frame corresponding to the geometric neighboring graph with $k$ nodes as $\text{g}_i\in(\text{O}_i,\textbf{E}_i,\textbf{W}_i)$.Then the voxel graph list is represented as $g=[\text{g}_1...\text{g}_T]$. In Figure \ref{fig2}, the voxel graph list maintains consistency with the event frames in the temporal dimension, and the three-dimensional geometric neighboring graph also compensates for the loss of intra-frame temporal information. 
Selecting voxels with with a higher count of points as nodes also emphasizes key local features.

\subsection{Feature Extraction}

Event frames can reflect the overall spatio-temporal information of the event stream but overlook the timestamp differences within a single frame, and they lack advantages in capturing local changes in the image. In contrast, voxel graph list maintain temporal associations within a frame, and the selection of graph nodes highlights areas with significant local changes, effectively representing the temporal information and local features within a frame. Consequently, we design a dual-branch network that extracts features from both image frames and voxel graph list. These features are then input into a feature fusion module, which concurrently perceives the global spatial and local spatio-temporal features.

\subsubsection{Event Frame Feature Extraction.}
In the branch of the event frame, we select ResBlock \cite{he2016deep}  as the core feature extractor to capture characteristics. ResBlock, centered around CNN, employs residual connections to circumvent gradient issues and is extensively utilized in feature extraction of image frames.  The event frame $f$ is initially subjected to convolution to acquire feature $F^f_{in}\in R^{T*C^f_{in}*H^f_{in}*W^f_{in}}$, which is then processed through ResBlock to yield feature $F^f_{out} \in R^{T*C^f_{out}*H^f_{out}*W^f_{out}}$ , as depicted below:
\begin{equation}
F^f_{out} = ResBlock(F^f_{in}).
\end{equation}

\subsubsection{Voxel Graph List Feature Extraction.}
In the branch of the voxel graph list, for the graph $g_i$
within the list, considering its node $o_i=((x_i,y_i,t_i),\mathbf{a_i})$ where $\mathbf{a_i}$ represents the feature vector, we utilize GMMConv \cite{monti2017geometric} and a GCNBlock with a residual structure to extract the spatio-temporal features of the collective neighboring graph. GMMConv employs a set of Gaussian kernels to perform convolution operations on each node in the graph and its neighbors. 
The GCNBlock is expressed as follows:
\begin{equation}
F^g_{res} = GMMConv(F^g_{in}),
\end{equation}
\begin{equation}
F^g_{out} = ELU(F^g_{res}+GMMConv(ELU(BN(F^g_{res})))),
\end{equation}
where $F^g_{in} \in R^{T*N^g_{in}*C^g_{in}}$ represents the initial features obtained from the voxel graph list $g$  after the convolution with the Gaussian Mixture Model, and $F^g_{out} \in R^{T*N^g_{out}*C^g_{out}}$ is the output feature after passing through the GCNBlock. $GMMConv()$ denotes the Gaussian Mixture Model convolution operation, which computes new node features based on node features $\mathbf{A_i}$, edge indices $\mathbf{E_i}$, and edge attributes $\mathbf{W_i}$.  $BN()$ stands for batch normalization, and $ELU()$ is the activation function applied to the output of the convolution.


\begin{figure*}[t!]
\centering
\scalebox{0.8}{\includegraphics[width=\textwidth]{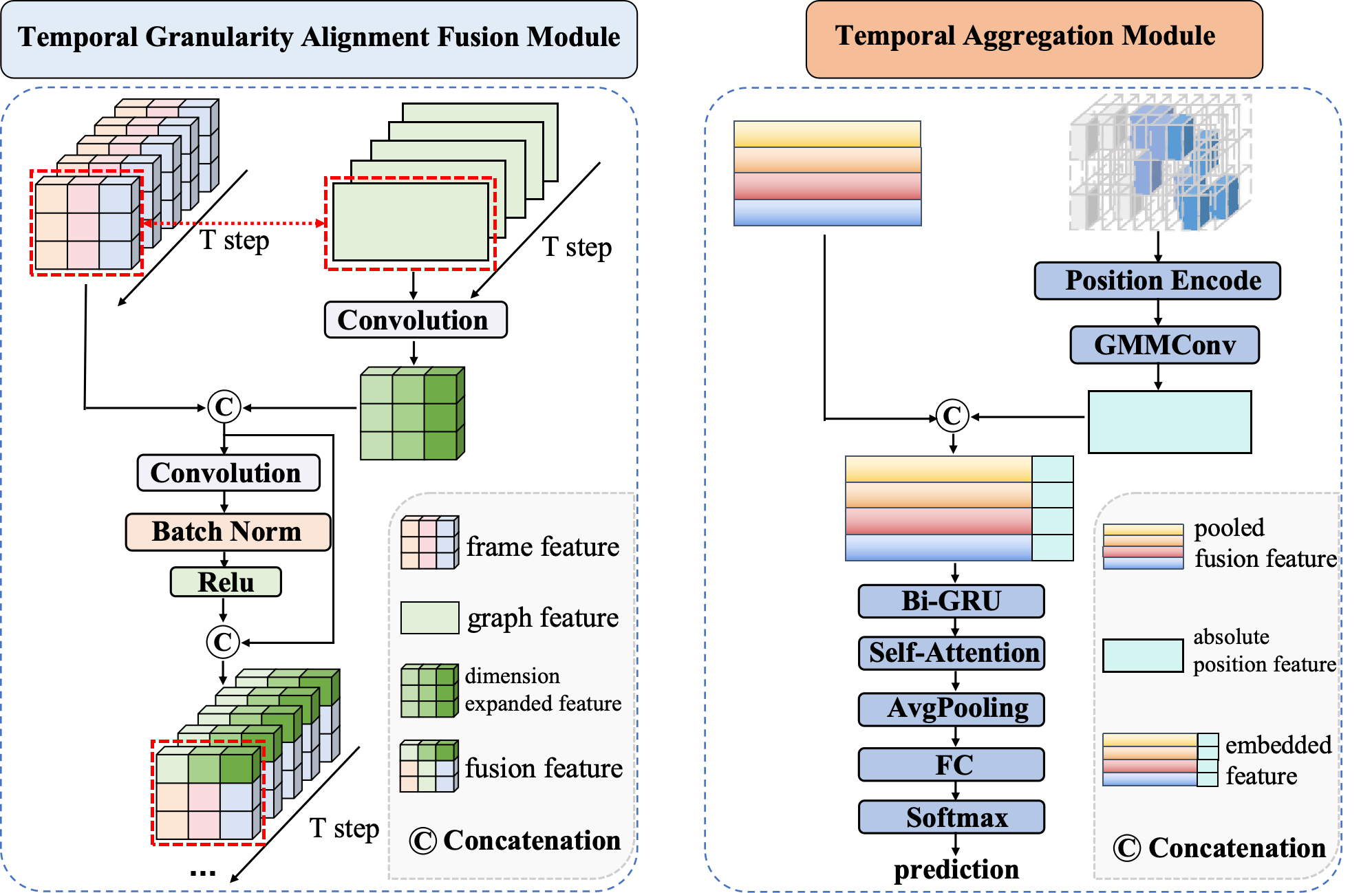}}
\caption{Illustrations of our designed two modules. (Left) The temporal granularity aligned fusion module. At each time step, the voxel graph node features are convolved to the same shape and then merged. The merged features achieve the fusion of spatio-temporal features through convolution and residual block. (Right) The temporal aggregation module. Extract features from the original voxel according to encoding, and then concatenate them with the sequence model, aggregating temporal information through Bi-GRU and Self-Attention layers.
} \label{fig3}
\end{figure*}

\subsection{Feature Fusion}
After obtaining the event frame and voxel graph list features, we design a fusion module centered around an iterative fusion mechanism in the fusion model for deeply integrating the event frame features and voxel grid features. This fusion model fully utilizes the complementary nature of these two types of features. Within the fusion module, based on the consistency of the features in the time dimension, we adopted temporal granularity alignment fusion approach. As illustrated in the Figure \ref{fig3}, for each event frame, a voxel grid from the same time segment is selected. The voxel graph list features are dimensionally expanded to adapt to the size of the event frame features, and then convolutional and residual modules are added to maintain the integrity of the information during the feature fusion process. The algorithm of the fusion module is as follows:
\begin{equation}
 F_{res} = Concat(F_{out}^f,Conv(F_{out}^g)),
\end{equation}
\begin{equation}
 F_{out}^{fus} = ELU(BN(Conv(F_{res}))) + F_{res}, 
\end{equation}
where $F_{out}^f$ and $F_{out}^g$ represent the output features of the previous module, serving as the input features of the fusion module.
$F_{out}^{fus}$ represents the output of the fusion module.

To fully explore the contextual associations and deep fine-grained information of the features while retaining the coarse-grained features presented by the frame features, we utilized an iterative strategy. In our configured N-layer (N=4 in our work) fusion model, each layer contains a fusion module. As shown in the feature extraction and fusion section of the Figure \ref{fig2}, the output features of the previous layer's fusion module are processed through a ResBlock for feature extraction, and re-entered into the fusion module together with the voxel grid features processed by the GMMBlock. Through multiple iterations, the features of the two branches are continuously fused, thereby realizing the role of voxel grid contextual features and local features in guiding the overall feature learning of the event frame. The complete iterative fusion block is represented as follows:

\begin{equation}
F^{fus}_i\!=\!FM(ResBlock(F^{fus}_{i-1}), GCNBlock(F^g_{i-1}))
\end{equation}
\begin{equation}
F_{out} = F^{fus}_N,
\end{equation}
where $F_i^{fus}$ denotes the output of the fusion module after the i-th iteration, $F_{out}$ denotes the fused features of the final output and $FM$ refers to FusionModule.

\subsection{Temporal Backend Network}
In this section, we propose a temporal back-end network based on Bi-GRU and Self-Attention with positional encoding. We first incorporate positional encoding using graph convolution to capture the absolute position information of nodes. This information is then concatenated with the spatio-temporal features obtained in the previous section. Subsequently, the Bi-GRU is employed to learn the temporal features of the feature sequences and contextual associations. Finally, the Self-Attention module is utilized to emphasize important temporal divisions.

Due to the properties of graph structures, graph convolution can only capture the relative positional relationships between nodes, corresponding to local spatio-temporal features in three-dimensional coordinates. As nodes' voxels are crucial local features, in order to capture the absolute position information of nodes, we apply positional encoding to the nodes before the temporal module. Specifically, we replace the polarity in voxel graph with the coordinates of nodes in the voxel grid as node features. This incorporation of node positions within the voxel grid allows the model to comprehend the spatial arrangement of nodes in the voxel grid, thus enhancing the discrimination ability of similar graph structures.

In the selection of temporal aggregation modules, similar to previous work \cite{tan2022multi}, we employ Bidirectional Gate Recurrent Units (Bi-GRU) to further aggregate temporal information, obtaining feature sequences containing temporal information. Through visual analysis, we observed variations in the number of event points across different temporal divisions. To ensure the model focuses on crucial sequences, we integrate a Self-Attention module to learn attention weights for each temporal division. These weights are then used to weight the feature sequences accordingly. The computation of the complete temporal back-end network is as follows:
\begin{equation}
Encode_{pos} = GMMConv(VoxelGraphList_{coor}),
\end{equation}
\begin{equation}
W_{att} = Softmax(\dfrac{QK^T}{\sqrt{d_k}})V,
\end{equation}
\begin{equation}
\scriptsize P = Softmax(\!FC(GAP_t(\!Bi(\!Cat(\!Encode_{pos},F_{out})*W_{att} \!)\!)\!)\!),
\end{equation}
where \textbf{$VoxelGraphList_{coor}$} represents voxel graph list with coordinates used as node features. $Encode_{pos}$ represents the positional encoding obtained. $Q$, $K$, and $V$ respectively represent the queries, keys, and values in Self-Attention.  $d_k$ denotes the dimensionality of the feature vectors, $W_{att}$ stands for the Self-Attention weights, and $P$ signifies the output probabilities for each word in the vocabulary.

\section{Experimental Evaluation}
\subsection{Evaluation Datasets}
\subsubsection{DVS-Lip.}
Acknowledging the significant contribution of Tan \cite{tan2022multi} \textit{et al.}  to event-based lip-reading tasks,  we utilize the DVS-Lip dataset they compiled, which is focused on lip-reading of words through an event camera. This dataset, gathered using event camera DAVIS346, is based on selecting words from the conventional camera-based lip-reading dataset LRW, including words that are often confused. It comprises event streams and intensity images for 100 words, with the lip region extracted within a 128*128 range. The validation of our model will be conducted using the DVS-Lip dataset, which will facilitate a comparison between our method and other approaches, highlighting our method's superiority.
\subsubsection{DVS128-Gait-Day.}To demonstrate the generalization capability of the models, we also conducted experiments on this dataset. The DVS128-Gait-Day\cite{wang2021event} dataset contains 4,000 gait event samples from 20 volunteers,who are asked to walk normally 100 times in front of a DVS128 sensor mounted on a tripod at approximately 90 degrees to the walking direction.
\begin{table*}[t!]
\centering
\begin{tabular}{c|c|c|c|c }

\hline
Model &  Representation  & Acc1(\%) & Acc2(\%) & Acc(\%) \\
\hline
Event Clouds \cite{wang2019space} &  {point clouds}  & 35.82 & 48.51 & 42.15 \\
EST \cite{gehrig2019end} &  {event spike tensor}  & 40.91 & 56.45 & 48.66\\
ACTION-Net \cite{wang2021action} & {video frame}  & 58.32 & 79.41 & 68.84\\
Martinez et al. \cite{martinez2020lipreading} & {video frame}   & 55.60 & 75.46 & 65.51\\
\hline
AGCN \cite{jiang2023point}& {(point,voxel)}   & 55.52 & 80.47 & 67.74 \\
GET \cite{peng2023get}& {group token}   & 58.96 & 80.82 & 69.80 \\
MSTP \cite{tan2022multi} & {event frame}  & 62.17 & 82.07 & 72.10 \\
\textbf{MTGA} & {\textbf{(event frame,voxel graph)}}  & \textbf{63.90} & \textbf{86.38} & \textbf{75.08} \\
\hline
\end{tabular}

\caption{Comparisons with existing event-based models and the state-of-the-art video-based models on the DVS-Lip test set. Acc1 represents the accuracy achieved on the first subset of the test data, Acc2 corresponds to the accuracy on the second subset, and Acc indicates the overall accuracy across the entire test dataset.}\label{tab1}
\end{table*}

\subsection{Experimental Results}
Since we utilized the same dataset and evaluation method as MSTP\cite{tan2022multi}, we selected MSTP for comparison. Additionally, we compared our method with some of the most advanced video or event-based action recognition and object recognition methods to demonstrate the superiority of our approach in the lip-reading task. 

Our experimental results, along with those of other methods, are presented in Table \ref{tab1}. The table illustrates that our Multi-view Temporal Granularity Aligned Aggregation significantly outperforms other action and object recognition methods in lip-reading tasks. This is primarily because the key to lip-reading is recognizing the syllable changes caused by lip movements. Recognition methods for objects generally lack the ability to capture local temporal features and struggle with lip-reading tasks that have a long time axis. Our method also surpasses the MSTP structure on the same dataset, indicating that our voxel graph list effectively compensates for the intra-frame temporal information lost during the temporal normalization of event frames. The structure of the voxel graph emphasizes the areas with a higher density of event points, adding local details to the overall features. Furthermore, the feature vectors representing the graph nodes effectively replenish some of the polarity information lost due to polarity weighted encoding.

In addition, we conducted experiments on the DVS128-Gait-Day dataset. As this is an event action stream dataset, we made a fair comparison with MSTP. The experimental results showed that under the same number of training epochs, MTGA achieved better performance. This proves that MTGA has broad applicability.

\subsection{Ablation Studies}

\subsubsection{Effects of Branch.}

We compared our fused model with individual branches in Table \ref{tab2}:(1) Only Event Frame Branch: this branch solely involves feature extraction for each frame of the event image. (2) Only Voxel Graph List Branch: this branch exclusively focuses on feature extraction from the temporally segmented voxel graph list. According to Table \ref{tab2}, both branches perform less effectively compared to the fused model. The Event Frame Branch, using CNN, loses temporal detail due to frame aggregation. The Voxel Graph List Branch, capturing temporal features with GMMConv, is less efficient and generalized than CNN. The data show that the accuracy of the Voxel Graph List Branch is only 70.03\%, which is 1.68\% lower than the Event Frame Branch, and the higher accuracy among the two branches is 3.37\% lower than the accuracy of the fused branch.
\begin{table}
\centering

\begin{tabular}{c|c|c }

\hline
Model &  Epoch & Acc(\%)  \\
\hline
MSTP & 10  & 90.28  \\
\textbf{MTGA} & \textbf{10} & \textbf{95.42} \\

\hline
\end{tabular}
\caption{Comparison results of different models on the DVS128-Gait-Day dataset.}\label{tab2}
\end{table}

\begin{table}
\centering

\begin{tabular}{c|c|c|c }

\hline
Branch &  Acc1(\%) & Acc2(\%) & Acc(\%) \\
\hline
Event Frame  & 60.05  & 83.81 & 71.71 \\
Voxel Graph List & 57.45 & 83.12 & 70.03\\
\textbf{Ours} &  \textbf{63.90} & \textbf{86.38} & \textbf{75.08}\\

\hline
\end{tabular}
\caption{Ablation study of adopting different branches.}\label{tab2}
\end{table}

\begin{table}
\centering

\begin{tabular}{c|c|c|c }

\hline
Fusion Method &  Acc1(\%) & Acc2(\%) & Acc(\%) \\
\hline
Weight  & 59.39  & 83.63 & 71.26 \\
Attention & 60.28 & 85.22 & 72.51 \\
\textbf{Ours} &  \textbf{63.90}  & \textbf{86.38} & \textbf{75.08} \\

\hline
\end{tabular}
\caption{Comparison of different fusion model.}\label{tab3}
\end{table}

\begin{table}
\centering

\begin{tabular}{c|c|c|c }

\hline
Model &  PEB  & AM & Acc(\%) \\
\hline
M1  & \ding{55}  & \ding{55} & 73.56 \\
M2 & \ding{51} & \ding{55} & 74.15\\
M3 & \ding{55} & \ding{51} & 74.87\\
\textbf{MTGA} &  \ding{51} & \ding{51} & \textbf{75.08}\\

\hline
\end{tabular}
\caption{Ablation study of adopting temperal aggregaion moudle.PEB refers to Position Embedding, and AM refers to Self-Attention Mechanism.}\label{tab4}
\end{table}


\begin{figure*}[t!]
\includegraphics[width=\textwidth]{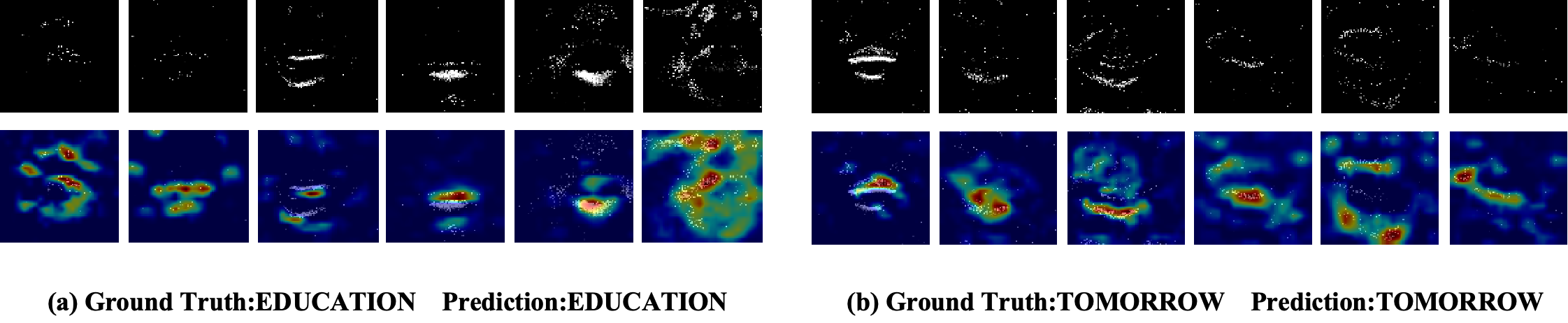}
\caption{Visualization of the saliency maps for words (a) “education” and (b) “tomorrow”.} \label{fig5}
\end{figure*}




\subsubsection{Effects of Fusion Methods.}
We further explored the impact of different fusion methods. We set up a comparison among three fusion approaches: (1) Overall feature fusion based on given weights, (2) Feature fusion based on the attention mechanism, (3) Fusion based on temporal granularity alignment.
The first method combined features linearly with fixed weights, a simple approach that lacks adaptability. The second employed an attention mechanism to dynamically allocate feature importance, enhancing adaptability. The third is the fusion based on temporal granularity alignment we introduced earlier, where the features of each frame are considered and fused individually, maintaining temporal details and dynamic changes.
Table \ref{tab3} displays our experimental results. It shows that the accuracy of the first fusion method is only 71.26\%. The accuracy of the second fusion method is 72.51\%. Although the attention mechanism can adaptively focus on the most useful features according to the context, in the lip-reading task where the front-end and back-end tasks are separated, the front-end feature fusion considers information within asynchronous time, which adversely affects the feature extraction at each timestep, leading to a decrease in accuracy. The experimental results further illustrate that our fusion method is the most effective in handling such time-series data.

\subsubsection{Effects of Temperal Aggregation Module.}
Our approach integrates a Bi-GRU and Self-Attention model with position encoding, enhancing the back-end network beyond the Bi-GRU used in \cite{tan2022multi}. Position encoding enriches the node positions absent in the front-end network. The Bi-GRU captures temporal dynamics and contextual links, followed by Self-Attention to weigh each time step.We conducted ablation experiments to verify the effectiveness by controlling whether to use position encoding and the Self-Attention module. The experimental results in Table \ref{tab4} demonstrate that compared to using only Bi-GRU (M1) as the back-end network, the accuracy is improved by  0.59\% with position encoding (M2),  1.31\% with the Self-Attention module (M3), and 1.52\% with both modules combined. It can be concluded that position encoding can indeed provide additional information, and Self-Attention can help our model focus on more important parts.

\subsection{Qualitative Analysis}
\subsubsection{Visualization of Frontend Network.}
To showcase the excellent effects of front-end feature extraction and fusion networks, we applied Grad-CAM \cite{selvaraju2017grad} to our model using samples from the DVS-Lip test dataset. The results from Grad-CAM vividly illuminate the visual saliency areas by computing gradients with respect to a specific class.

As shown in Figure \ref{fig5}, we present two examples to demonstrate the effectiveness of our model. With 60 bins, we selected one event frame image every 10 bins, where the top row shows the input original images, and the bottom row displays the overlaid images with heatmaps generated by applying Grad-CAM to the last network layer. It is observable that, after the fusion of the two branches, our model focuses on the most crucial local information in the images,
while also attributing certain weight to the scattered points around the edges, representing the global information.

\section{Conclusion}
In this paper, we propose a model named MTGA, wherein the feature of a large number of event data are extracted from multiple views.
Through our designed Temporal Granularity aligned Fusion Module, the fused features simultaneously possess global spatial and local spatio-temporal information, addressing the issue of feature loss caused by previous single-frame aggregating methods.
Experiments on the DVS-Lip dataset validate the superiority of our model. From the results, we mainly conclude that: (1) Each branch of the model has the ability to capture different features, and our fusion module effectively integrates them.; (2) Our fusion approach has a competitive advantage over other methods; (3) Our supplementary position encoding and Self-Attention effectively improve the aggregation effect of the back-end network.
In the future, we intend to integrate other event views and further explore fusion strategies to enable the model to comprehensively reflect the features. Additionally, we also aims to enhance the model’s generalization performance to effectively handle other event recognition tasks.

\section*{Acknowledgments}
This work is supported in part by the National National Key Research and Development Program of China (No. 2022YFF0712300), the National Natural Science Foundation of China (Grant No. U23A20318, 62276195 and 62376200), the Science and Technology Major Project of Hubei Province (Grant No. 2024BAB046) and the Innovative Research Group Project of Hubei Province (Grant No. 2024AFA017). The numerical calculations in this paper have been done on the supercomputing system in the Supercomputing Center of Wuhan University.

\bibliography{aaai25}

\end{document}